\documentclass[11pt,a4paper]{article}
\usepackage[hyperref]{acl2021}
\usepackage{times}
\usepackage{latexsym}
\usepackage{xcolor,soul}
\usepackage{multicol}

\usepackage{todonotes}

\usepackage{microtype}
\usepackage{enumerate}
\usepackage{amssymb}
\usepackage{amsmath}
\usepackage{latexsym}
\usepackage{url}
\usepackage{mathtools}
\usepackage{multirow}
\usepackage{xspace}
\usepackage{comment}
\usepackage{flowchart}
\usetikzlibrary{arrows}
\usepackage{booktabs}
\usepackage{subcaption}
\usepackage{hyperref}
\usepackage{enumitem}
\usepackage{multirow}
\usepackage{tikz}
\usepackage{pgfplots}

\usepackage{float}
\usepackage{multicol}
\restylefloat{table}
\usepackage{amsmath}
\usepackage{graphicx}
\usepackage{color}
 
\usepackage{comment}
\usepackage{enumerate}
\usepackage{enumitem}

\usepackage{float}
\restylefloat{table}
\usepackage{placeins}

\usepackage[nameinlink]{cleveref}
\crefformat{section}{\S#2#1#3} 
\crefformat{subsection}{\S#2#1#3}
\crefformat{subsubsection}{\S#2#1#3}

\usepackage{xcolor}
\usepackage[linesnumbered, algo2e, vlined, ruled]{algorithm2e}

\SetCommentSty{mycommfont}

\DeclareMathAlphabet{\pazocal}{OMS}{zplm}{m}{n}
\DeclareMathAlphabet{\pazocal}{OMS}{zplm}{m}{n}

\usepackage{cleveref}

\crefformat{section}{\S#2#1#3} 
\crefformat{subsection}{\S#2#1#3}
\crefformat{subsubsection}{\S#2#1#3}

\aclfinalcopy 

\newcommand{\eg}{{\em e.g.,}\xspace}

\newcommand{\Ni}{{(i)}~}
\newcommand{\Nii}{{(ii)}~}
\newcommand{\Niii}{{(iii)}~}

\newcommand{\Na}{({\em a})~}
\newcommand{\Nb}{({\em b})~}

\newcommand{\augvic}{{\sc{AugVic}}}

\title{{\augvic}: Exploiting BiText Vicinity for Low-Resource NMT}

\author{Tasnim Mohiuddin\thanks{\ \ Equal contribution}$^{\ \ \P}$, M Saiful Bari $^{*\P}$, \and Shafiq Joty$^\P$$^\dagger$ \\
$^\P$Nanyang Technological University, Singapore \\
$^\dagger$Salesforce Research Asia, Singapore \\
{\tt \{mohi0004, bari0001, srjoty\}@ntu.edu.sg}
\\}

\date{}

\begin{document}
\maketitle

\begin{abstract}

The success of Neural Machine Translation (NMT) largely depends on the availability of large bitext training corpora. Due to the lack of such large corpora in low-resource language pairs,  NMT systems often exhibit poor performance. Extra relevant monolingual data often helps, but acquiring it could be quite expensive, especially for low-resource languages. Moreover, domain mismatch between bitext (train/test) and monolingual data might degrade the performance. To alleviate such issues, we propose \augvic, a novel data augmentation framework for low-resource NMT which exploits the vicinal samples of the given bitext without using any extra monolingual data explicitly. It can diversify the in-domain bitext data with finer level control. Through extensive experiments on four low-resource language pairs comprising data from different domains, we have shown that our method is comparable to the traditional back-translation that uses extra in-domain monolingual data. When we combine the synthetic parallel data generated from \augvic\ with the ones from the extra monolingual data, we achieve further improvements. We show that \augvic\ helps to attenuate the discrepancies between relevant and distant-domain monolingual data in traditional back-translation. To understand the contributions of different components of \augvic, we perform an in-depth framework analysis.
\end{abstract}







\section{Introduction}
\label{sec:intro}

Neural Machine Transaltion (NMT) has shown impressive performance in high-resource settings, even claiming to achieve parity with human professional translators \cite{Hassan2018AchievingHP,popel2020transforming}. Most successful NMT systems have billions of parameters \cite{lepikhin2021gshard}. They generally work well only when a good amount of parallel training data is available and perform poorly in low-resource conditions \cite{koehn-knowles-2017-six,guzman-etal-2019-flores}. However, majority of the languages are low-resourced despite being used by large portion of world population. Hence, improving low-resource MT quality has been of great interests to the MT researchers.  



There have been several attempts to extend the success of NMT in high-resource settings to low-resource language pairs that have a relatively small amount of available parallel data. Most of these methods mainly focus on leveraging extra monolingual data through back-translation \cite{sennrich-etal-2016-improving} and self-training  \cite{He2020Revisiting}, or translation knowledge transfer through parallel data involving other assisting language pairs \cite{firat-etal-2016-multi,firat-etal-2016-zero,johnson-etal-2017-googles,neubig-hu-2018-rapid}.\footnote{See \cite{survey-mnmt} for a survey of the later.} Large scale pre-training is another recent trend to utilize large monolingual data for NMT \cite{liu2020multilingual}. However, very few work has considered low-resource NMT without using auxiliary data or other pivot languages.




In the presence of a sufficient amount of in-domain monolingual data, back-translation (BT) has proved to be quite successful \cite{edunov-etal-2018-understanding}. In this approach, a reverse intermediate model is trained on the original parallel data, which is later used to generate synthetic parallel data by translating sentences from target-side monolingual data into the source language. 
However, when there are scarcity of in-domain data which indeed a common situation in many low-resource settings, the success of BT may be limited \cite{chen-etal-2019-facebook}. 

Another understudied problem with BT is the issue with \emph{domain mismatch} \cite{edunov-etal-2020-evaluation}. To elaborate, let us consider two scenarios: \Ni the training and testing data come from the same or relevant domains (\eg News), and \Nii  the test domain (News) is different from the training domain (\eg Subtitles). In the former case, we can foresee two problems. First, if we use out-of-domain monolingual data which is abundant, it might misguide the model and move it  far away from the actual test distribution. Second, even if the monolingual data is from a domain similar to that of the training/testing data, there might be differences in topics, modality, style, etc., which might induce noise.

 
For the latter scenario, even if the monolingual data comes from the similar domain as the test data (News), the corresponding (reverse) translations will be noisy as the intermediate model would be trained on a different domain (Subtitles). Consequently, these noisy pseudo-parallel data will induce noise during training and might cause the model to perform worse \cite{wang-etal-2018-switchout}. On the other hand, using in-domain (Subtitles) monolingual data in back-translation will not give enough diversity to cover the test domain (News). 

\begin{figure*}[t!]
\centering
\scalebox{1.0}{
  \includegraphics[width=1\linewidth,trim=1 1 1 1,clip]{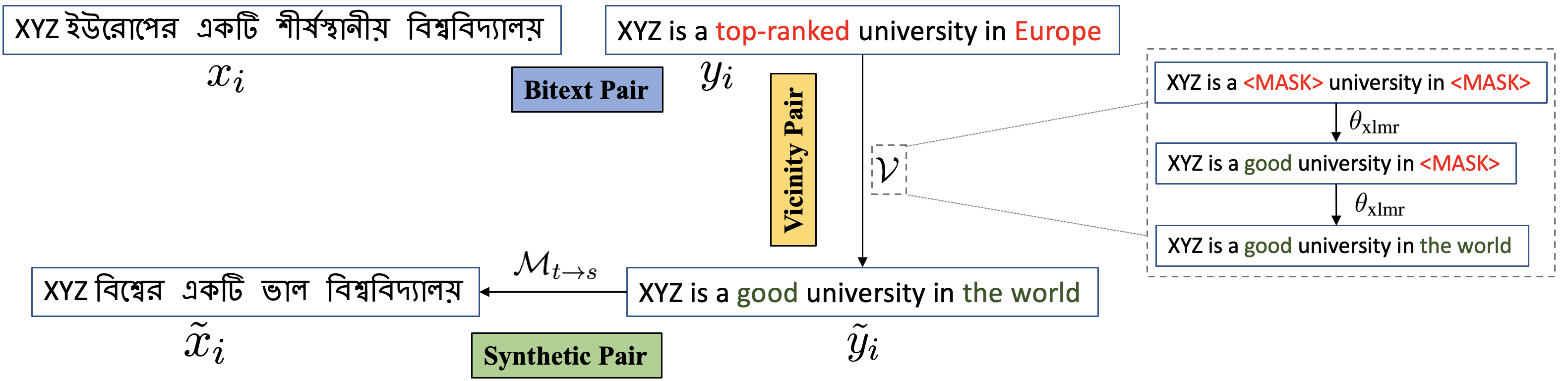}}
\caption{Illustration of \augvic\ steps for Bengali-to-English translation system. Here $(x_i, y_i)$ is the original bitext pair, $\Tilde{y}_i$ is a vicinal sample of $y_i$, and $(\Tilde{x}_i, \Tilde{y}_i)$ is a synthetic pair where $\Tilde{x}_i$ is generated by a reverse intermediate translation system $\mathcal{M}_{t \rightarrow s}$. Right side of the figure shows the successive steps of vicinal sample generation. }
\label{fig:illust-example}
\end{figure*}

In this work, inspired by the Vicinal Risk Minimization principle \cite{VRM_NIPS2000}, we propose \augvic, a novel method to \textbf{aug}ment \textbf{vic}inal samples around the bitext distribution. Instead of using extra monolingual data, \augvic\ aims to leverage the vicinal samples of the original bitext, thereby enlarging the support of the training bitext distribution to improve model generalization. The main advantage is that the resulting  distribution remains close to the original distribution and can be controlled at a finer level (\Cref{fig:illust-example}).

With the goal of training a source-to-target NMT system, \augvic\ augments vicinal samples in the target language. The vicinal samples are generated by predicting the masked tokens of a target bitext sentence using a pretrained large-scale language model. To generate synthetic bitext data from these augmented vicinal samples through a reverse intermediate (target-to-source) model, we propose two different methods: the first one is based on the traditional BT, while the second one leverages the original source sentence as a guide. Finally, we train the source-to-target model by combining the original parallel data with the synthetic bitext.

In order to demonstrate the effectiveness and robustness of \augvic, we conduct extensive experiments on four low-resource language pairs comprising data from different domains. Our results show significant improvements over the bitext baselines with 2.76 BLEU gains on an average on eight different translation tasks without using any extra monolingual data. \augvic\ also complements traditional BT with additive gains when extra monolingual data is used. We also show \augvic's efficacy in bridging the gap between in-domain and out-of-domain performance in traditional back-translation with monolingual data. We carried out an ablation study to understand the contribution of the diversity factor in our proposed framework. We open-source our framework at \href{https://ntunlpsg.github.io/project/augvic/}{https://ntunlpsg.github.io/project/augvic/}. 


 
  
  
   

\section{Related Work}



Two lines of studies are relevant to our work.

\paragraph{Low-resource NMT}

Although the main focus of investigation and improvement in NMT has been in high-resource settings, there has been a recent surge of interest in low-resource MT. However, achieving satisfactory performance in low-resource settings turns out to be challenging for NMT systems \cite{koehn-knowles-2017-six}. Recent research has mainly focused on creating and cleaning parallel \cite{en-tam, gz0b-5p24-18} and comparable data \cite{TIEDEMANN12.463}, utilizing bilingual lexicon induction \cite{MUSE,artetxe2018acl,mohiuddin-joty-2019-revisiting,mohi-et-al-journal-word-tr,mohiuddin-etal-2020-lnmap}, fine-grained hyperparameter tuning \cite{sennrich-zhang-2019-revisiting}, and {using other language pairs as pivot \cite{ijcai2017-555,kim-etal-2019-pivot}.}

Another avenue of research follows multilingual translation, where translation knowledge from high-resource language pairs are exploited by training a single NMT system on a mix of high-resource and low-resource language pairs \cite{firat-etal-2016-multi,firat-etal-2016-zero,kocmi-bojar-2018-trivial,DBLP:journals/corr/abs-1802-05368,neubig-hu-2018-rapid,guzman-etal-2019-flores}. 
\citet{zoph-etal-2016-transfer} proposed a variant where they pretrain NMT system on a high-resource language pair before finetuning on a target low-resource language pair.

\paragraph{Data Augmentation for NMT}

Till now, one of the most successful data augmentation strategies in NMT is back-translation (BT) \cite{sennrich-etal-2016-improving, hoang-etal-2018-iterative}, which exploits target-side monolingual data. \citet{edunov-etal-2018-understanding} investigated BT extensively and scaled the method to millions of target-side monolingual sentences. \citet{caswell-etal-2019-tagged} explored the role of noise in noised-BT and proposed to use a tag for back-translated source sentences. Besides BT, self-training is another data augmentation strategy for NMT which leverages source-side monolingual data \cite{He2020Revisiting}. Large scale multilingual pre-training followed by bitext fine-tuning is a recent trend to utilize monolingual data for NMT, which is shown to be beneficial \cite{M4, liu2020multilingual, Zhu2020Incorporating,lepikhin2021gshard}.


Apart from using extra monolingual data, \citet{XieWLLNJN17} show that data noising is an effective regularization method for NMT, while \citet{wu-etal-2019-exploiting} use noised training. In low-resource settings, \citet{fadaee-etal-2017-data} augment bitext by replacing a common word with a low-frequency word in the target sentence, and change its corresponding word in the source sentence to improve the translation quality of rare words. \citet{wang-etal-2018-switchout} propose an unsupervised data augmentation method for NMT by replacing words in both source and target sentences based on hamming distance. \citet{gao-etal-2019-soft} propose a method that replaces words with a weighted combination of semantically similar words. Recently, \citet{NguyenJWA20} propose an in-domain augmentation method by diversifying the available bitext data using multiple forward and backward models. In their follow-up work \cite{nguyen2020multiagent}, they extend the idea to unsupervised MT (UMT) using a cross-model distillation method, where one UMT model's synthetic output is used as input for another UMT model. 


\paragraph{Summary}
Most of the previous work on improving BT involve either training iteratively or combining BT with self-training using monolingual data blindly {without noticing the distributional differences between the monolingual and bitext data}. In contrast, in \augvic\ we systematically parameterize the generation of new training samples from the original parallel data. Moreover, the combination of our augmented vicinal samples with monolingual data makes the NMT models more robust and attenuates the prevailing distributional gap.

\section{Method}

Let $s$ and $t$ denote the source and target languages respectively, and $\mathcal{D} = \{(x_i, y_i)\}_{i=1}^N$ denote the bitext training corpus containing $N$ sentence pairs with $x_i$ and $y_i$ coming from $s$ and $t$ languages, respectively. Also, let $\mathcal{M}_{s \rightarrow t}$ is an NMT model that can translate sentences from $s$ to $t$, and $\mathcal{D}_{\text{mono}}^t = \{y_j\}_{j=1}^{M}$ denote the monolingual corpus in the target language $t$ containing $M$ sentences.


\subsection{Traditional Back-Translation}
\label{subsec:traditional-bt} 
Traditional back-translation \cite{sennrich-etal-2016-improving} leverages the target-side monolingual corpus. With the aim to train a source-to-target model $\mathcal{M}_{s \rightarrow t}$, it first trains a reverse intermediate model $\mathcal{M}_{t \rightarrow s}$ using the given bitext $\mathcal{D}$, and use it to translate the extra target-side monolingual data $\mathcal{D}_{\text{mono}}^t$ into source language. This yields a synthetic bitext corpus $\mathcal{D}_{\text{syn}} = \{ \mathcal{M}_{t \rightarrow s}(y_j), y_j)\}_{j=1}^M$. Then a final model $\mathcal{M}_{s \rightarrow t}$ is trained on $\{ \mathcal{D} \cup \mathcal{D}_{\text{syn}}\}  $ usually by upsampling $\mathcal{D}$ to keep the original and synthetic bitext pairs to a certain ratio (generally 1:1).

\subsection{\augvic: Exploiting Bitext Vicinity}

For low-resource languages, the amount of available parallel data is limited, hindering training of a good MT system. Moreover, the target language pairs can be quite different (\eg\ morphologically, topic distribution) from the high-resource ones, making the translation task more difficult \cite{chen-etal-2019-facebook}. Also, acquiring large and relevant monolingual corpora in the target language is difficult in low-resource settings and can be quite expensive. The domain mismatch between the monolingual and bitext data is another issue with the traditional back-translation as mentioned in \S \ref{sec:intro}. 



With the aim to improve model generalization, the core idea of \augvic\ is to leverage the \emph{vicinal} samples of the given bitext rather than using extra monolingual data. The addition of bitext vicinity also alleviates the domain mismatch issue since the augmented data distribution does not change much from the original bitext distribution. Figure \ref{fig:illust-example} shows an illustrative example of \augvic, which works in three basic steps to train a model:


\begin{enumerate}[label=(\roman*)]
    \item Generate vicinal samples $\Tilde{y}_i$ of the target sentences ($y_i$) in the bitext data $\mathcal{D}$.
    \item Produce source-side translations $\Tilde{x}_i$ of the vicinal samples to generate synthetic bitext $\Tilde{\mathcal{D}}$.
    \item Train the final source-to-target MT model $\mathcal{M}_{s \rightarrow t}$ using $\{ \mathcal{D} \cup \Tilde{\mathcal{D}}\}$.
\end{enumerate}

\augvic, however, is not mutually exclusive to the traditional back-translation and can be used together when relevant monolingual data is available. In the following, we describe how each of these steps are operationalized with NMT models.

\subsubsection{Generation of Vicinal Samples}
\label{subsec:gen-vicinal}

We first generate vicinal samples for each eligible target sentence $y_i$ in the bitext $\mathcal{D} = \{(x_i, y_i)\}_{i=1}^N$. Let $\mathcal{V}(\Tilde{y}_i|{y_i})$ denote the vicinity distribution around $y_i$, we create a corpus of vicinal samples as:  
\begin{eqnarray}
     \Tilde{y}_i \sim \pazocal{V}(\Tilde{y}_i| y_i)
\end{eqnarray}
We generate vicinal samples for sentences having lengths between 3 and 100, and $\pazocal{V}$ can be modeled with existing syntactic and semantic alternation methods like language model (LM) augmentation \cite{kobayashi,conditional_bert,aug_bert,bari-et-al-uxla}, paraphrase generation {\cite{li-etal-2018-paraphrase}}, constrained summarization \cite{summary_loop}, and similar sentence retrieval \cite{du2020selftraining}. Most of these methods are supervised requiring extra annotations. Instead, in \augvic, we adopt an unsupervised LM augmentation, which makes the framework more robust and flexible to use. Specifically, we use a pretrained XLM-R masked LM \cite{xlmr} parameterized by $\theta_{\text{xlmr}}$ as our vicinal model. Thus, the vicinity distribution is defined as $\pazocal{V}(\Tilde{y}_i| y_i, \theta_{\text{xlmr}})$. 

{Note that we treat the vicinal model as an external entity, which is not trained/fine-tuned. This disjoint characteristic gives our framework the flexibility to replace $\theta_{\text{xlmr}}$ even with a better monolingual LM for a specific target language, which in turn makes \augvic\ extendable to utilize stronger LMs that may come in the future.}

In a masked LM, one can mask out a token at any position and ask the model to predict at that position. For a meaningful and informed augmentation, we mask out the tokens \textit{successively} (one at a time) up to a required number {determined by a diversity ratio, $\rho \in (0,1)$}. For a sentence of length $\ell$, the successive augmentation can generate at most $(2^{\ell}-1) \times k$ vicinal samples, where $k$ is the number of output tokens chosen for each masked position. We use $k=1$, and pick the one with the highest probability ensuring that it does not match the original token at the masked position. The diversity ratio ($\rho$) controls how much diverse the vicinal samples can be from the original sentence, and is selected using one of the following two ways:

 


\begin{itemize}[leftmargin=*]
\item \textbf{Fixed diversity ratio~~} Here we use a fixed value for $\rho$, and select $t = {\ell \times \rho}$ tokens to mask out.  We then generate new vicinity samples by predicting new tokens in those masked positions. 

\item \textbf{Dynamic diversity ratio~~} Instead of using a fixed value, in this approach we set the diversity ratio dynamically taking the sentence length into consideration. This allows finer-level control for diversification --- the longer the sentence is, the smaller should its diversification ratio be. The intuition is that for long sentences, a larger value of $\rho$ will produce vicinal samples which will be far away from the original sample. Specifically, we use the following piece-wise function to find the number of tokens to mask out dynamically:
\begin{eqnarray}
     t = \begin{cases} 
          \max(\ell \times a, t_{\text{min}}) & ; \text{if } \ell \leq 20 \\
          \min(\frac{\ell}{h} \times b, t_{\text{max}}) & ; \text{otherwise}
   \end{cases}
   \label{eq:diversity}
\end{eqnarray}
where $t_{\text{min}}$ and $t_{\text{max}}$ are hyperparameters and represent the minimum and maximum number of tokens to be replaced by the masked LM. The other hyperparameters $a$, $b$, and $h$ play the same role as the diversity ratio $\rho$.
\end{itemize} 

Since we predict tokens for replacement one at a time, we can make the prediction in any of the permutation order of $t$. So, the maximum number of possible augmentation for a sentence of length $\ell$ is  $\gamma = \binom{\ell}{t} \times {t}!$. We perform \textit{stochastic sampling} from the distribution of $\gamma$ to select ${N^{\prime}}$ vicinal samples. We have added an analysis on the effect of diversity ratio $\rho$ in \augvic\ in \S \ref{subsec:diversity}.

\subsubsection{Generation of Synthetic Bitext Data}

Our objective is to train a source-to-target MT model $\mathcal{M}_{s \rightarrow t}$. So far, we have the bitext $\mathcal{D} = \{(x_i, y_i)\}_{i=1}^N$ and target-side monolingual data $\Tilde{\mathcal{D}}^t  = \{\Tilde{y}_j\}_{j=1}^{N^{\prime}}$ which are vicinal to the original target in $\mathcal{D}$. We need a reverse intermediate target-to-source MT model $\mathcal{M}_{t \rightarrow s}$ to translate $\Tilde{y}_j$ into $\Tilde{x}_j$, which will give us the synthetic bitext data $\Tilde{\mathcal{D}}$. For this, we experiment with two different models. 

\begin{description}[leftmargin=0pt]
\item [\Na Pure Back-Translation (PBT)]
\label{subsub:pbt}
This is similar to back-translation (\S \ref{subsec:traditional-bt}), where we first train the reverse MT model $\mathcal{M}_{t \rightarrow s}$ using the given bitext $\mathcal{D}$. We then use $\mathcal{M}_{t \rightarrow s}$ to translate the target-side vicinal samples $\Tilde{y}_j \sim \Tilde{\mathcal{D}}^t$ into $\Tilde{x}_j$. This gives a synthetic bitext  $\Tilde{\mathcal{D}} = \{(\Tilde{x}_j, \Tilde{y}_j)\}_{j=1}^{N^{\prime}}$. We use the Transformer architecture \cite{NIPS2017_3f5ee243} as our reverse intermediate NMT model $\mathcal{M}_{t \rightarrow s}$.

\begin{figure}[t!]
\centering
\scalebox{0.75}{
  \includegraphics[width=1.3\linewidth,trim=1 1 1 1,clip]{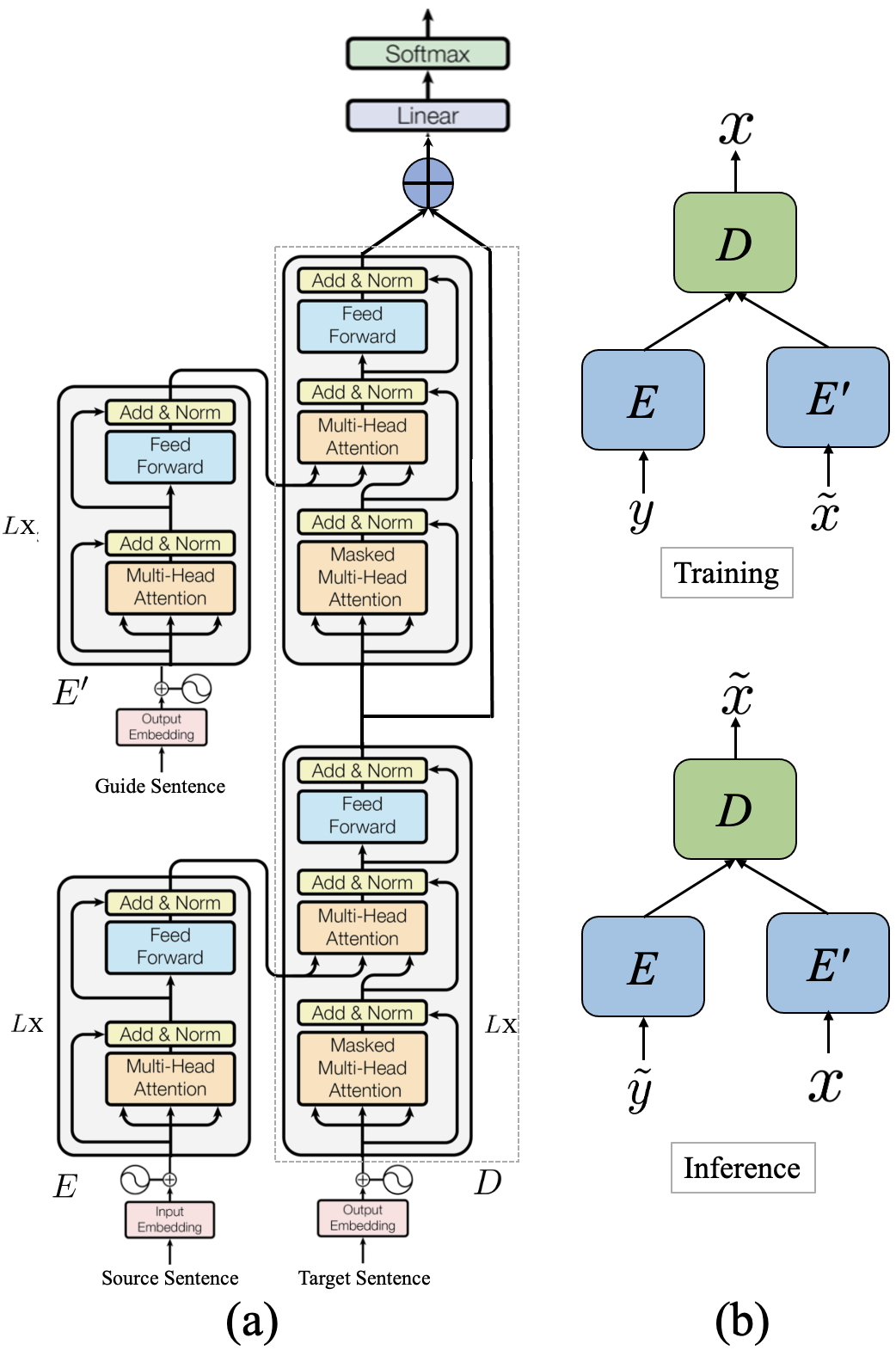}}
\caption{(a) Our proposed model for guided back-translation; (b) its training and inference method. 
}
\label{fig:gbt}
\end{figure}

\item [\Nb Guided Back-Translation (GBT)]
\label{subsub:gbt}
In the illustrative example (\Cref{fig:illust-example}), we can identify three kinds of pairs: \Ni the bitext $(x_i, y_i)$, \Nii the vicinal $(y_i, \Tilde{y_i})$, and \Niii the synthetic pair $(\Tilde{x_i}, \Tilde{y_i})$. Here, $y_i$ is the original translation of source sentence $x_i$ and $\Tilde{y_i}$ is the vicinal sample, which can be seen as a perturbation of ${y_i}$. Hence, we can assume that $\Tilde{x_i}$ will also be similar to (perturbed) $x_i$. Our goal is to leverage this extra relational knowledge to improve the translation quality of $\Tilde{x_i}$ when generating the synthetic bitext $\Tilde{\mathcal{D}}$. Specifically, we use the original source $x_i$ as a guide for generating the synthetic translation $\Tilde{x_i}$ of the target-side vicinal sample $\Tilde{y_i}$. 
\begin{eqnarray}
     \Tilde{x_i} = \mathcal{M}_{t \rightarrow s}(\Tilde{y_i}| x_i) 
\end{eqnarray}
For this, we propose a model based on the Transformer architecture which has two encoders - one for the source sentence ($E$) and another for the guide sentence ($E'$), and a decoder ($D$) (\Cref{fig:gbt}). We use the same architecture with the exception that now we have two identical encoders ($E$ and $E'$). Both the encoders have a stack of $L$ layers, while the decoder has $(L+1)$ layers.

\textit{Training \& Inference:} We train this model with a dataset of triplets containing $(y$, $\tilde{x}$, $x)$, where $(x, y)$ comes from the original bitext and $\tilde{x}$ is a vicinal sample of $x$ to guide the decoder in generating $x$. Each of the first $L$ layers of the decoder performs cross-attention on $E(y)$ resulting in decoder states  $D^{\tiny{(L)}}(x_{<t}|y)$ at time step $t$, while the final decoder layer attends on $E'({\tilde{x}})$ resulting in a second set of decoder states  $D^{\tiny{(L+1)}}(x_{<t}|y, {\tilde{x}})$. The two sets of decoder states are then interpolated by taking a convex combination before passing it to a linear layer followed by the Softmax token prediction.
\begin{equation}
    \lambda  D^{\tiny{(L)}}(x_{<t}|y) + (1-\lambda) D^{\tiny{(L+1)}}(x_{<t}|y, {\tilde{x}}) 
    \label{eq:gbd}
\end{equation}
\noindent where $\lambda$ is a hyperparameter that controls the relative contributions from the two encoders, $E(y)$ and $E'({\tilde{x}})$, in generating $x$ by the decoder $D$.

To generate the synthetic bitext $\tilde{D}$, we need to translate $\tilde{y}$, which will be guided by $x$. So during \textit{inference}, we feed $\tilde{y}$ to $E$ and $x$ to $E^{\prime}$ to autoregressively generate $\tilde{x}$ with beam search decoding.






\end{description}


\subsubsection{Training of the Final Model} 

We combine the original bitext $\mathcal{D}$ and the synthetic bitext $\Tilde{\mathcal{D}}$ generated from the previous step to train our final source-to-target model $\mathcal{M}_{s \rightarrow t}$. We use the standard Transformer as our final model.

\begin{table*}[t]
\centering
\scalebox{0.9}{
\begin{tabular}{l|c|c|c}
\toprule
{\textbf{Pair}} & \textbf{Data-Source} & \textbf{Train \& Dev}  & \textbf{Test} \\ 
\midrule
En-Bn      &  \citet{gz0b-5p24-18} & Mixed  &  Mixed    \\
En-Ta      & \citet{en-tam} &  News, Bible, Cinema    & {News, Bible, Cinema}    \\ 
\midrule
En-Ne     &   \citet{guzman-etal-2019-flores} &   Bible, GV, PTB, Ubuntu  &  Wikipedia  \\ 
En-Si     &   \citet{guzman-etal-2019-flores} &   Opens subtitles, Ubuntu  &  Wikipedia  \\ 
\bottomrule
\end{tabular}}
\caption{Sources and domains of the datasets.}
\label{tab:domain}
\end{table*}

\section{Experimental Setup}

\subsection{Datasets and Evaluation Metrics} \label{subsec:data}

We conduct experiments on four low-resource language pairs: English (En) to/from Bangla (Bn), Tamil (Ta), Nepalese (Ne), and Sinhala (Si). Table \ref{tab:domain} presents the source of the collected datasets and their domains for each language pair. 

Even though the En-Bn dataset size is relatively small {($\sim$ 72K pairs)}, the quality of the bitext is rich, and it covers a diverse set of domains including literature, journalistic texts, instructive texts, administrative texts, and texts treating external communication. Here the distributions in train and test splits are about the same. For En-Ta, the train and test domains are similar, mostly coming from the news ({$\sim$ 66.43\%}). For En-Ne and En-Si, we use the datasets from \cite{guzman-etal-2019-flores}, where the train and test domains are different. Although these two datasets are comparatively larger ($\sim$ 600K pairs each), the quality of the bitext is poor, {requiring further cleaning and deduplication}. 

Table \ref{tab:stats} presents the dataset statistics after deduplication where the last column specifies the number of augmented data by our method \augvic\ (\S \ref{subsec:gen-vicinal}). For a fair comparison with the traditional back-translation, we experiment with the same amount of target-side monolingual data from three domains: news, wiki, and gnome. We collected and cleaned News, Wiki, and Gnome datasets from News-crawl, Wiki-dumps, and Gnome localization guide, respectively. For some languages, the amount of specific domain monolingual data is limited, where we added additional monolingual data of that language from Common Crawl.




Following previous work \cite{guzman-etal-2019-flores, NguyenJWA20}, we report the tokenized BLEU \cite{papineni-etal-2002-bleu} when translating from English to other languages, and detokenized SacreBLEU \cite{post-2018-call} when translating from other languages to English for all our experiments,. 

\begin{table}[H]
\centering
\scalebox{0.7}{
\begin{tabular}{l|c|c|c||c}
\toprule
{\textbf{Pair}} & \textbf{Train} & \textbf{Dev}  & \textbf{Test} & \textbf{Augmented} (\augvic/Mono) \\ 
\midrule
En-Bn      &  70,854  & 500      &  500 & $\approx$ 460K    \\
En-Ta      & 166,851 &  1000    & 2000  & $\approx$ 1300K   \\ 
\midrule
En-Ne     &   234,514  &   2559  &  2835 & $\approx$ 1500K  \\ 
En-Si     &  571,213  &  2898   & 2766  & $\approx$ 1500K  \\ 
\bottomrule
\end{tabular}
}
\caption{Dataset statistics after deduplication.}
\label{tab:stats}
\end{table}

\subsection{Baselines}
We compare \augvic\ with the following baselines:

\begin{description}[leftmargin=0pt,itemsep=-0.0em]
\item [\Ni {Bitext baseline}] is the model trained with the bitext given with the dataset.
\item [\Nii {Upsample baseline}] Here we upsample the bitext to the same amount of \augvic's data.
\item [\Niii {Diversification baseline}] \citet{NguyenJWA20} diversifies the original parallel data by using the predictions of multiple forward and backward NMT models. Then they merge the augmented data with the original bitext on which the final NMT model is trained. Their method is directly comparable to \augvic, as both methods diversify the original bitext, but in different ways. 

\end{description}



\begin{table*}[t!]
\centering
\small
\scalebox{1.0}{
\begin{tabular}{ll|cc|cc|cc|cc}
\toprule
\textbf{Setting} & \textbf{Data} & \multicolumn{2}{c}{\textbf{En-Bn}}  &        \multicolumn{2}{c}{\textbf{En-Ta}} &\multicolumn{2}{c}{\textbf{En-Ne}}& \multicolumn{2}{c}{\textbf{En-Si}}
\\

& & \textbf{$\rightarrow$} & \textbf{$\leftarrow$} & \textbf{$\rightarrow$} & \textbf{$\leftarrow$} & 
\textbf{$\rightarrow$} & \textbf{$\leftarrow$} & 
\textbf{$\rightarrow$} & \textbf{$\leftarrow$} 
\\   
\toprule
\multirow{ 2}{*}{Baseline} &
 {Bitext} & 13.21 & 21.18 & 11.58 & 26.29 & 4.59 & 8.34 & 1.96 & 7.45 \\

& \quad $\times$ Upsample &  16.59 & 25.51 & 12.15 & 27.71 & 4.16 & 7.79 & 1.81 & 6.93 \\

\midrule
\multirow{ 2}{*}{Diversification} & \quad + \citet{NguyenJWA20} & 17.54 & 26.11 & 12.74 & 28.54 & 5.7 & 8.9 & 2.2 & 8.2 \\
& \quad + \augvic & \textbf{18.03} & \textbf{26.96} & \textbf{12.93}  & \textbf{28.68}  & \textbf{6.47} & \textbf{10.65} & \textbf{3.66} & \textbf{9.27} \\
\midrule[0.8pt]

\multirow{ 4}{*}{Extra mono. data} & \quad + BT-Mono (News) & 18.81 & 27.11 & 13.51 & 29.38 & 6.44 & 12.48 & 3.56 &  11.75 \\

& \quad + BT-Mono (Wiki) & 18.52 & 26.33 & 13.23  & 29.01  & 6.91 & 13.02 & 3.91 & 11.86 \\
\cmidrule[0.2pt]{2-10}

& \quad + \augvic + BT-Mono (News) &  19.98 & 28.14 & 13.87 & \textbf{30.15} & 6.80 & 13.12 & 4.94 & 11.89 \\

& \quad + \augvic + BT-Mono (Wiki) &  \textbf{20.39} & \textbf{28.48} & \textbf{13.89} & 30.14 & \textbf{7.27} & \textbf{13.52} & \textbf{5.24} & \textbf{12.09} \\


\bottomrule
\end{tabular}}
\caption{{Detokenized Sacre-BLEU} scores for \{Bn, Ta, Ne, Si\} $\rightarrow$ En and tokenized BLEU fro En $\rightarrow$ \{Bn, Ta, Ne, Si\}. ``BT-Mono'' stands for traditional back-translation with extra target-side monolingual data (\S \ref{subsec:traditional-bt}). } 
\label{tab:main-results} 
\end{table*}

\subsection{Model Settings}



{We use the Transformer \cite{NIPS2017_3f5ee243} implementation in Fairseq \cite{ott-etal-2019-fairseq}. We follow the basic  architectural settings from \cite{guzman-etal-2019-flores}, which establishes some standards for low-resource MT. For low-resource ``Bitext baseline'', they use  a smaller (5-layer) Transformer architecture as the dataset is small, while for larger datasets (\eg\ with additional synthetic data) they use a bigger (6-layer) model.\footnote{\href{https://github.com/facebookresearch/flores/}{https://github.com/facebookresearch/flores/}} To keep the architecture the same in the respective rows (Table \ref{tab:main-results}), we use a 6-layer model for ``Upsample baseline'' and 5-layer for ``Bitext baseline''. More specifically, for datasets with less than a million bitext pairs, we use an architecture with 5 encoder and 5 decoder layers, where the number of attention heads, embedding dimension, and inner-layer dimension are respectively 8, 512, and 2048. Otherwise, we use a larger Transformer architecture with 6 encoder and 6 decoder layers with the number of attention heads, embedding dimension, and inner-layer dimension of 16, 1024, and 4096, respectively.}


After deduplication, we tokenize non-English data using the Indic NLP Library.\footnote{\href{https://github.com/anoopkunchukuttan/indic_nlp_library}{https://github.com/anoopkunchukuttan/indic\_nlp\_library}} We use the sentencepeiece library\footnote{\href{https://github.com/google/sentencepiece}{https://github.com/google/sentencepiece}} to learn the joint Byte-Pair-Encoding (BPE) of size 5000 symbols for each of the language pair over the raw English and tokenized non-English bitext training data.

{We tuned the hyper-parameters $a$, $b$, $h$, $t_{min}$, $t_{max}$ in Eq.  \ref{eq:diversity} and $\lambda$ in Eq. \ref{eq:gbd} by small-scale experiments on the validation-sets. We found $a = 0.5$, $b=2.5$, $h=10$, $t_{min}=1$, and $t_{max}=20$ work better. We tuned $\lambda$ within the range of  $0.5$ to $0.9$. In general, we observe that for smaller sentences (length $<=$ 20), 50-60\% successive-token-replacement works better while for longer sentences (length $>$ 20), 20-30\% token-replacement performs better.}

{Following \citet{guzman-etal-2019-flores}, we train all the models upto a maximum epoch of 100 with early-stopping enabled based on the validation loss. We use the beam-search-decoding for inference. All the reported results for \augvic\ use dynamic diversity ratio for generating vicinal samples unless otherwise specified.}

\section{Results and Analysis}
In this section, we present our results and the analysis of our proposed methods.

\subsection{Comparison with Bitext \& Diversification}

Table \ref{tab:main-results} presents the BLEU scores on the eight translation tasks. First, we compare our model \augvic\ with the model trained on the original parallel data (Bitext). \augvic\ consistently improves the performance over all the tested language pairs, gaining about $+2.76$ BLEU scores on average. 
Specifically, \augvic\ achieves the absolute improvements of 4.28, 5.78, 1.35, 2.39, 1.88, 2.31, 1.70, and 1.82 over the Bitext for En-Bn, Bn-En, En-Ta, Ta-En, En-Ne, Ne-En, En-Si, and Si-En, respectively.

For a fair comparison, in another experiment, we upsample the bitext data to make it similar to the amount of \augvic 's data. From the \emph{Upsample} results {(with a 6-layer architecture)} reported in Table \ref{tab:main-results}, we see that even though it increases the BLEU scores for En to/from \{Bn, Ta\}, it has negative impacts on En to/from \{Ne, Si\} where it degrades the performance. Overall, \augvic\ achieves 1.75 BLEU score improvements on an average over the Upsample baseline.

The comparison with the diversification strategy proposed by \citet{NguyenJWA20} reveals that \augvic\ outperforms their method by 0.84 BLEU scores on average. To be specific, {our method gets 0.49, 0.85, 0.19, 0.14, 0.77, 1.75, 1.46, and 1.07 absolute BLEU improvements over their approach for En-Bn, Bn-En, En-Ta, Ta-En, En-Ne, Ne-En, En-Si, and Si-En, respectively.} 

{The data diversification method of \citet{NguyenJWA20} relies heavily on the performance of base models (Bitext). From Table \ref{tab:main-results}, we see that the performance of base models are poor for En to/from \{Ne, Si\}, which impacts their augmented data generation process (diversification). However, the better performance of \augvic\ in those languages indicates that vicinal samples generated in our method are more diverse with better quality and less prone to the noise in base models.}





\begin{table*}[t!]
\centering
\small
\scalebox{1.0}{
\begin{tabular}{l|cc|cc|cc|cc}
\toprule
\textbf{Interm. } & \multicolumn{2}{c}{\textbf{En-Bn}}  &        \multicolumn{2}{c}{\textbf{En-Ta}} &\multicolumn{2}{c}{\textbf{En-Ne}}& \multicolumn{2}{c}{\textbf{En-Si}}
\\

\textbf{BT system} & \textbf{$\rightarrow$} & \textbf{$\leftarrow$} & \textbf{$\rightarrow$} & \textbf{$\leftarrow$} & 
\textbf{$\rightarrow$} & \textbf{$\leftarrow$} & 
\textbf{$\rightarrow$} & \textbf{$\leftarrow$} 
\\   
\toprule
{Pure BT} & 18.03 & 26.96 & 12.93  & 28.68  & \textbf{6.47} & \textbf{10.65} & \textbf{3.66} & \textbf{9.27}  \\

{Guided BT} & \textbf{18.18} & \textbf{27.35} & \textbf{13.17} & \textbf{29.05} & 4.81 & 8.62 & 2.16 & 7.71  \\

\bottomrule
\end{tabular}}
\caption{Comparison between two \textbf{interm}ediate reverse back-translation \textbf{(BT)} systems in \augvic.} 
\label{tab:pbt-gbt} 
\end{table*}

\subsection{Vicinal Samples with Extra Relevant Monolingual Data}

We further explore the performance of \augvic\ by experimenting with the traditional back-translation method (\S \ref{subsec:traditional-bt}) using the same amount of monolingual data. To perceive the variability, we choose to experiment with extra monolingual data from two \emph{relevant} but different sources - newscrawl (BT-Mono (News)) and Wikipedia (BT-Mono (Wiki)). From the results in Table \ref{tab:main-results}, we see that standard back-translation improves the scores in both cases, proving that extra relevant monolingual data helps for low-resource MT significantly. 

To understand the exclusivity of the vicinal samples of \augvic\ from the external related monolingual data, we perform another set of experiments where we added both the \augvic's augmented data with the extra  monolingual data and trained along with the Bitext data. From Table \ref{tab:main-results}, we see that the combination of datasets improves the BLEU scores by 1.02 and 0.73 on average on the two relevant data sources (News and Wiki). From this, we can conclude that vicinal samples of \augvic\ make the NMT models more robust in the presence of the relevant monolingual data and can be used together when available.

\subsection{Pure vs. Guided: Which One is Better?}

For all the results of \augvic\ presented in Table \ref{tab:main-results}, we use the pure back-translation (BT) method (\S \ref{subsub:pbt}(a)) as the reverse intermediate model. We compare the performance of the guided BT (\S \ref{subsub:gbt}(b)) with the pure BT  method  as the reverse intermediate model in Table \ref{tab:pbt-gbt}. From the results, we observe that the guided BT achieves better results in En$\leftrightarrow$ \{Bn, Ta\}, while the pure BT achieves better in En$\leftrightarrow$ \{Ne, Si\} translation tasks. 

We investigated why the guided BT performed poorly in En$\leftrightarrow$ \{Ne, Si\} tasks, and found that compared to the En-Bn and En-Ta bitexts, the original bitexts of En-Ne and En-Si languages are very noisy (\eg\ bad sentence segmentation, code-mix data), which propagates further noise while using the target translation as a guide for translating the vicinal samples. The diminishing results while upsampling in these two languages (Table \ref{tab:main-results}) supports this claim. From these results, we can say that the better the original bitext quality is, the better the synthetic bitext will be for the guided BT.


\subsection{\augvic\ with Relevant and Distant-domain Monolingual Data}

To verify how traditional back-translation and \augvic\  perform with with monolingual data from related vs. distant domains, we perform another set of experiments on En to/from \{Bn, Ta\}. For both the language pairs (\Cref{subsec:data}), \textit{News} can (roughly) be considered as relevant compared to \textit{gnome},\footnote{\href{http://opus.nlpl.eu/GNOME.php}{http://opus.nlpl.eu/GNOME.php}} which can be considered as distant domain. We use \textit{pure BT} as the intermediate reverse back-translation system for generating synthetic data in \augvic\ in this set of experiments.


\begingroup
\begin{table}[t!]
\centering
\small
\scalebox{0.85}{
\begin{tabular}{ll|cc|cc}
\toprule
\textbf{BT-mono} & \textbf{Data} & \multicolumn{2}{c}{\textbf{En-Bn}}  &        \multicolumn{2}{c}{\textbf{En-Ta}}
\\

\textbf{Domain}  &  & \textbf{$\rightarrow$} & \textbf{$\leftarrow$} & \textbf{$\rightarrow$} & \textbf{$\leftarrow$}  
\\   
\toprule
& Bitext & 13.21 & 21.18 & 11.58 & 26.29 \\
\midrule
\emph{News} & + BT   & 18.81 & 27.11 & 13.51 & 29.38  \\

{(relevant)} & + \augvic + BT    & 19.98 & 28.14 & 13.87 & 30.15  \\

\midrule

\emph{gnome} & + BT   & 17.14 & 26.05 & 12.55 & 27.91  \\

{(distant)} & + \augvic + BT  & 18.86 & 27.56 & 13.59 & 29.89 \\

\bottomrule
\end{tabular}}
\caption{Effect of relevant and distant domain monolingual data in back-translation with \augvic. We use \textit{News} as ``relevant''  and \textit{gnome} as ``distant'' domain.} 
\label{tab:out-domain} 
\end{table}
\endgroup



From Table \ref{tab:out-domain}, we see that traditional back-translation (+ BT) improves the BLEU scores over the Bitext by 4.14 and 2.85 on average for relevant- and distant-domain monolingual data, respectively, yielding higher gains for relevant domain, as expected. The addition of vicinal data by \augvic\ (+ \augvic + BT) further improves the scores in both cases; interestingly, the relative improvements are higher in the distant-domain case. Specifically, the average BLEU score improvements over Bitext for relevant- and distant-domain data with \augvic+BT are 4.97 and 4.41, respectively. Comparing this with BT only, the BLEU score difference between relevant and distant domains has been reduced from 1.29 to 0.56. This indicates that \augvic\ helps to bridge the domain gap between relevant and distant-domain distributions in traditional BT with monolingual data. 

{In principle, for vicinal samples, the synthetic-pair generation capability of the reverse intermediate target-to-source MT model should be better than generating from an arbitrary monolingual data as it could be a distant distribution compared to the bitext. Judging by the amount of diverse data used for training the language model, we can safely assume that it is a diverse knowledge source \cite{conneau-etal-2020-unsupervised} compared to the training bitext samples. Data that performs well on the reverse intermediate target-to-source MT system can be extrapolated from the knowledge-base as vicinal-distribution with the controlled diversity ratio function (Eq. \ref{eq:diversity}). Moreover, to achieve more diversity, the use of multiple different language models is also compatible in \augvic.}




\subsection{Effect of Diversity Ratio in \augvic}
\label{subsec:diversity}

For monolingual data, it could be challenging to identify domain discrepancy with the training/testing bitext data, and there is no parameter {in the traditional BT method} to control this distributional mismatch. However, in \augvic\ we can control the distributional drift of the generated vicinal samples from the original training distribution by varying the diversity ratio $\rho$.

{Theoretically, it is possible to sample the same distribution using dynamic and static diversity. However, dynamic diversity is more flexible to perform hyperparameter-tuning and to prevent potential outliers. The term $l/h$ in Eq. \ref{eq:diversity} represents pseudo-segmentation ($h$ segments) of a large sentence of length $l$, and $b$  represents the same intuition as $\rho$. Apart from these, $t_{min}$ and $t_{max}$ prevents irregular-samples: (i) $t_{min}$ ensures that there should be at least some changes in the augmented sample, (ii) $t_{max}$ makes sure that the generated-samples from LM do not diverge too much from the vicinity.}

To understand the effect of the diversity ratio in \augvic, we perform another set of experiments. We choose to use En to/from \{Bn, Ne\} for this experiments, where we selected at most two vicinal samples from each of the target sentence in original bitext.  We investigate the effect of both \textit{dynamic} and \textit{fixed} diversity ratio in \augvic's vicinal sample generation (\S \ref{subsec:gen-vicinal}). For fixed diversity ratio we use $\rho$ values 0.1, 0.3, 0.5, and 0.8, while for dynamic diversity ratio we use $a=0.5, b=2.5,$ and $h=10$ for controlling the diversity. 

We present these experimental results in Table \ref{tab:diversity-ratio}, from where we see that the dynamic diversity ratio performs better in three out of four tasks. For the fixed diversity ratio, we see the variation in results for different values of $\rho$. In all the four tasks, the diversity ratio $\rho=0.8$ gives the least scores. On average, we get the better results with $\rho=\{0.3, 0.5\}$. These experiments suggest that higher diversity values may induce noise and lower diversity values may not diversify the data enough to benefit the final NMT model.


\begingroup
\begin{table}[t!]
\centering
\small
\scalebox{1.1}{
\begin{tabular}{l|cc|cc}
\toprule
\textbf{\augvic } & \multicolumn{2}{c}{\textbf{En-Bn}}  &        \multicolumn{2}{c}{\textbf{En-Ne}}
\\
\textbf{diversity ratio}
& \textbf{$\rightarrow$} & \textbf{$\leftarrow$} & \textbf{$\rightarrow$} & \textbf{$\leftarrow$}  
\\   
\toprule
Dynamic   & \textbf{17.69} & \textbf{26.61} & \textbf{6.21} & 10.25  \\

\midrule
Fixed \\
\quad $\rho=0.1$ & 17.34 & 25.98 & 5.98 & 10.03  \\

\quad $\rho=0.3$ & 17.52 & 26.19 & 6.19 & 10.36  \\

\quad $\rho=0.5$ & 17.48 & 26.49 & 6.05 & \textbf{10.38}  \\

\quad $\rho=0.8$ & 17.19 & 25.01 & 5.82 & 9.89 \\

\bottomrule
\end{tabular}}
\caption{Effect of diversity ratio $\rho$ while generating vicinal samples in \augvic\ (\S \ref{subsec:gen-vicinal}). } 
\label{tab:diversity-ratio} 
\end{table}
\endgroup



\section{Conclusion}
We have presented an in-domain data augmentation framework \augvic\  by exploiting the bitext vicinity for low-resource NMT. Our method generates vicinal samples by diversifying sentences of the target language in the bitext in a novel way. It is simple yet effective and can be quite useful when extra in-domain monolingual
data is limited.

Extensive experiments with four low-resource language pairs comprising data from different domains show the efficacy of \augvic. Our method is not only comparable with traditional back-translation with in-domain monolingual data, it also makes the NMT models more robust in the presence of relevant monolingual data. Moreover, it bridges the distributional gap for out-of-domain monolingual data when using together.

\bibliographystyle{acl_natbib}
\bibliography{anthology,acl2021}

\appendix
\clearpage
\section{Appendix}

\subsection{Reproducibility Settings}
\begin{itemize}
    \item Computing infrastructure - Linux machine with Tesla V100-SXM2-16GB GPU
    \item PyTorch version: 1.4.0
    \item CUDA version: 10.2
    \item cuDNN version: 7.6
    \item Number of model parameters - 
    \begin{itemize}
        \item Base Model: 39340032
        \item Larger Model: 181481472
        \item Guided BT Model: 90039296
    \end{itemize}
\end{itemize}

\subsection{Optimal Hyperparameters}

\begin{table}[H]
\begin{center}
\scalebox{0.75}{
\begin{tabular}{l|c}
\hline 
\bf Hyperparameter & \bf Value \\  
\hline
Transformer Layers & 5\\
Emb. dim & 512\\
FFN dim & 2048\\
Attention heads & 8\\
Share-all-embeddings & \texttt{True}\\
Dropout & 0.3 \\
Label-smoothing & 0.2 \\
Warmup-updates & 4000 \\ 
Warmup-init-lr & 1e-7 \\
Learning rate & 0.003 \\ 
Min-lr  & 1e-9 \\
Optimizer & \texttt{adam} \\ 
Adam-betas & \texttt{(0.9, 0.98)} \\
Lr-scheduler & \texttt{inverse-sqrt} \\
Criterion & \texttt{\small{label-smooth-cross-entropy}} \\
\hline
\end{tabular}
}
\end{center}
\caption{\label{autoenc-hyperparameters} Optimal hyper-parameter settings for base model.}
\end{table}

\begin{table}[H]
\begin{center}
\scalebox{0.75}{
\begin{tabular}{l|c}
\hline 
\bf Hyperparameter & \bf Value \\  
\hline
Transformer Layers & 6\\
Emb. dim & 1024\\
FFN dim & 4096\\
Attention heads & 16\\
Share-all-embeddings & \texttt{True}\\
Dropout & 0.1 \\
Label-smoothing & 0.2 \\
Warmup-updates & 4000 \\ 
Warmup-init-lr & 1e-7 \\
Learning rate & 0.001 \\ 
Min-lr  & 1e-9 \\
Optimizer & \texttt{adam} \\ 
Adam-betas & \texttt{(0.9, 0.98)} \\
Lr-scheduler & \texttt{inverse-sqrt} \\
Criterion & \texttt{\small{label-smooth-cross-entropy}} \\
\hline
\end{tabular}
}
\end{center}
\caption{\label{autoenc-hyperparameters} Optimal hyper-parameter settings for large model.}
\end{table}

\end{document}